\documentclass[letterpaper, 10 pt, conference]{ieeeconf}  

\IEEEoverridecommandlockouts                              
\usepackage{graphics} 
\usepackage{epsfig} 
\usepackage{mathptmx} 
\usepackage{times} 
\usepackage{amsmath} 
\usepackage{amssymb}  
\usepackage{algorithm,algorithmic}
\usepackage{scalerel}
\usepackage[dvipsnames]{xcolor}
\usepackage{multirow} 
\usepackage{makecell} 
\usepackage{cite} 

\title{\LARGE \bf
A Robotic System for Implant Modification in Single-stage Cranioplasty
}

\author{Shuya Liu$^{1}$, Wei-Lun Huang$^{2}$, Chad Gordon$^{3}$ and Mehran Armand$^{4}$
\thanks{$^{1}$ Shuya Liu is with the Department of Mechanical Engineering, Johns Hopkins University, Baltimore, MD. {\tt\small jsliu@jhu.edu}}
\thanks{$^{2}$ Weilun Huang is with the Department of Computer Science, Johns Hopkins University, Baltimore, MD. {\tt\small wl.huang@jhu.edu }}
\thanks{$^{3}$ Chad Gordon is with the Department of Plastic and Reconstructive Surgery, Johns Hopkins School of Medicine, Baltimore, MD. {\tt\small cgordon@jhmi.edu}}
\thanks{$^{4}$ Mehran Armand is with the Department of Orthopaedic Surgery, Mechanical Engineering, and Computer Science, Johns Hopkins University, Baltimore, MD. {\tt\small marmand2@jhu.edu}}
}

\begin{document}

\maketitle
\thispagestyle{empty}
\pagestyle{empty}

\begin{abstract}
Craniomaxillofacial reconstruction with patient-specific customized craniofacial implants (CCIs) is most commonly performed for large-sized skeletal defects. Because the exact size of skull resection may not be known prior to the surgery, in the single-stage cranioplasty, an oversized CCI is prefabricated and resized intraoperatively with a manual-cutting process provided by a surgeon. The manual resizing, however, may be inaccurate and significantly add to the operating time. This paper introduces a fast and non-contact approach for intraoperatively determining the exact contour of the skull resection and automatically resizing the implant to fit the resection area. Our approach includes four steps: First, we acquire a patient’s defect information using a handheld 3D scanner. Second, the scanned defect is aligned to the CCI by registering the scanned defect to the preoperative CT model. Third, a cutting toolpath is generated from the scanned defect model by extracting the resection contour. Lastly, a cutting robot resizes the oversized CCI to fit the resection area. To evaluate the resizing performance of our method, we generated six different resection shapes for the cutting experiments. We compared the performance of our method to the performance of surgeon’s manual resizing and an existing technique that collects the defect contour with an optical tracking system. The results show that our proposed method improves the resizing accuracy by 56\% compared to the surgeon’s manual modification and 42\% compared to the optical tracking method.
\end{abstract}

\section{Introduction}


\begin{figure}[t!]
\centerline{\includegraphics[width=1.0\linewidth]{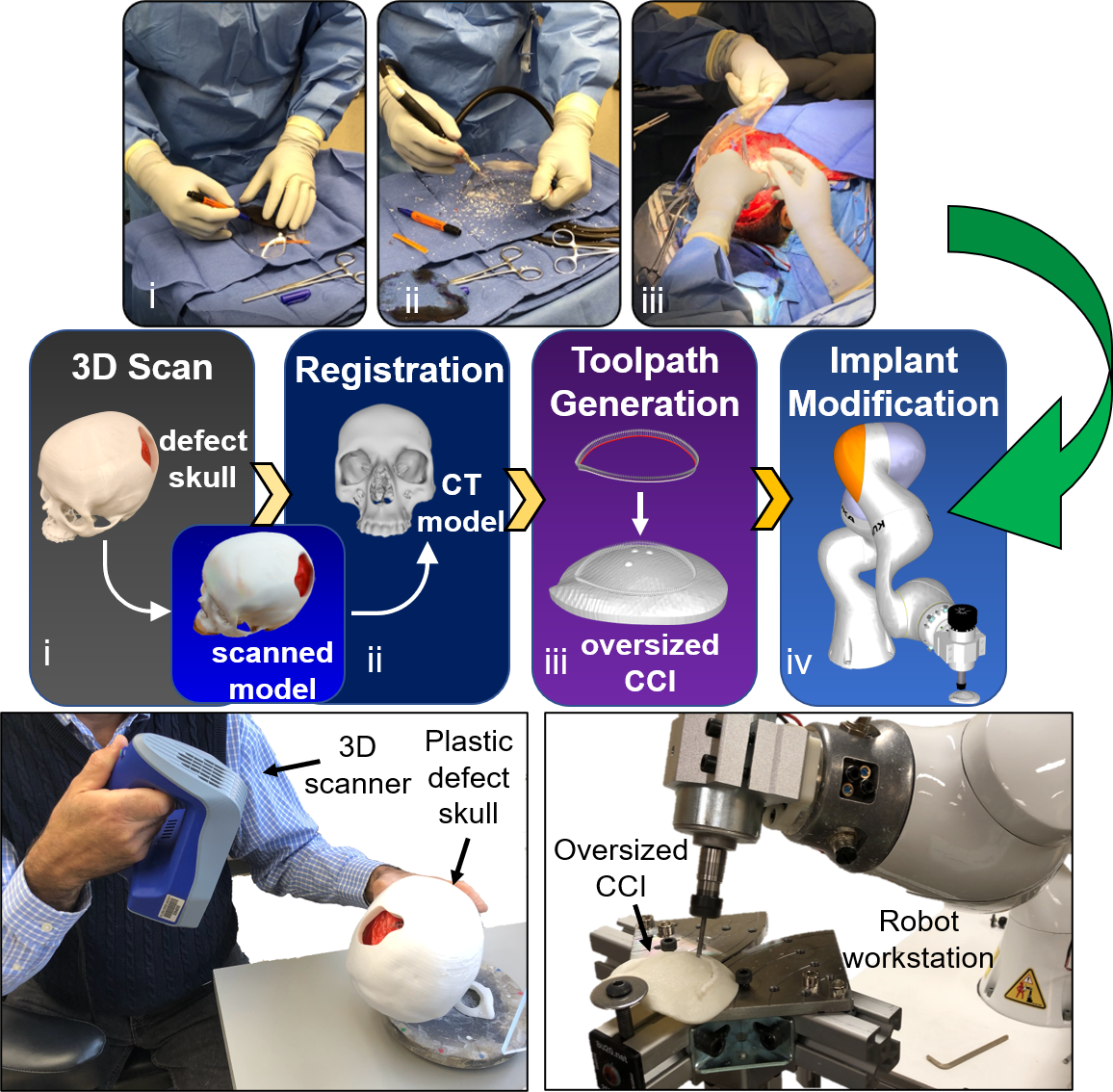}}
\caption{Top: A clinical example of the single-stage cranioplasty with a prefabricated oversized CCI: i) the surgeon marks the defect contour on the CCI. ii) the implant is manually modified by a surgical cutter. iii) the resized CCI is fit to the skull defect. Middle: The workflow of robotic single-stage cranioplasty: i) 3D scanning generates a scanned model of the defect skull. ii) the scanned model is registered to the preoperative CT model. iii) a cutting toolpath is generated. iv) a cutting robot modifies the implant. Bottom: Left: The plastic defect skull is scanned by a handheld 3D scanner. Right: robot resizes the oversized implant.}
\label{fig-overview}
\end{figure}


Cranioplasty is a procedure to treat cranial defects due to trauma, injury, or neurosurgical procedures for brain tumors, aneurysms, or epilepsy \cite{aydin2011cranioplasty}. Conventional cranioplasty is a two-stage process that repairs skull deformities in a delayed operation \cite{pasick2019adult}. Such process requires the skull to be partially removed from the patient, who then has to wait for the design and fabrication of the replacing implant for three to four weeks. In contrast, single-stage cranioplasty aims to restore aesthetic appearance immediately following craniectomy within one single operation, therefore, decreasing operative times and speeding up the patient's recovery \cite{berli2015immediate}. 

Several approaches are utilized to generate CCIs. The molding technique has been applied to form CCIs in the operating room \cite{fathi2008cost}. This method requires injecting liquid bio-compatible materials such as poly-methyl-methacrylate directly into the defect of the skull or a molding template generated using the autologous bones. However, molding a CCI directly on the defect may release an exothermic reaction damaging nearby tissue\cite{marbacher2012intraoperative}. Moreover, autologous bones cannot always be used to create a negative imprint and are limited in their ability to eliminate the discontinuities between the boundaries. Another commonly used approach is a cutting guide, in which a customized implant and a cutting guide are prefabricated, and the surgeon resects the patient's skull following the cutting guide so that the skull defect can be immediately closed with the prefabricated implant \cite{eufinger1998single, gerbino2013single}. In addition, some other groups considered using optical navigation systems to achieve planned resections. Although these methods are capable of repairing skull deformities within one operation, they do not consider the possibilities for intraoperative plan changes, limiting neurosurgeons' flexibility in reaching specific regions of the brain \cite{jalbert2014one, dodier2020single}.

In practice, cranioplasty replaces the skull defect with an alloplastic implant instead of using a patient’s autologous skull-bone \cite{shah2014materials}. Although additive manufacturing has been widely used to fabricate CCIs \cite{kim2012customized, park2016cranioplasty, morales2018cranioplasty}, it is not realistic to bring 3D printers into the operating room to intraoperatively print CCIs due to its long material building process and sterilization requirement. Therefore, the subtractive approach with prefabricated and pre-sterilized oversized CCIs is more practical and cost-effective. In \cite{berli2015immediate, wolff2018adult}, a clinical approach using prefabricated oversized CCIs in single-stage cranioplasty is presented. This approach requires a surgeon to intraoperatively modify an oversized CCI by manually resizing it, which is often poor in accuracy and time-consuming.

Computer-assisted single-stage cranioplasty provides a method to help surgeons better visualize the defect contour by directly projecting the defect contour on an oversized CCI. This method utilizes an optical tracking system to collect data points of the defect contour\cite{murphy2015computer}. However, this system is difficult to set up and requires line-of-sight. Moreover, the planar projection from a fixed configuration may not be suitable for implants with complex structures. To address the above-mentioned problems, we previously developed a portable projection mapping device that tracks surgical instruments and projects a 3D defect contour onto the implant in real-time from any angle without information loss \cite{liu2020portable}. Although this approach improves the accuracy of projection mapping for medical augmented reality, it can only collect one data point per frame with a digitizing instrument. Therefore, this approach takes longer to collect sufficient data points for registration.  

Recent advances in 3D scanning technologies provide new venues for extending the application of medical robots in the operating room \cite{haleem20193d}. 3D scanning generates high-precision 3D models of real-world objects that can be recognized within a robot's workspace to achieve specific autonomous tasks. Different from the optical tracking system, a 3D scanner can collect thousands of data points per frame without contacting the object. The use of a 3D scanner for skull defect reconstruction can simplify and expedite the identification of the defect's contour.

In this paper, we present a novel system for generating precise CCIs for patients in single-stage cranioplasty. The system consists of a 3D scanner and a cutting robotic arm. The 3D scanning technique enables fast registration and generation of cutting toolpaths. The cutting robotic arm provides stable and accurate performance compared to the conventional manual cutting approach by surgeons and an existing method using an optical tracking system.

The contributions of this work include:
\begin{itemize}
\item We proposed a fast and non-contact approach for acquiring defect contour information using a handheld 3D scanner.
\item We developed an algorithm for generating cutting toolpaths by extracting defect contours.
\item We integrated a robotic system for automated implant modification.
\end{itemize}

\section{Method}\label{overview}
We developed a robotic system for resizing oversized CCIs during intraoperative operation. The system consists of a handheld Artec Space Spider 3D scanner (up to 0.1 mm resolution) and a KUKA LBR iiwa 7 R800 robotic arm (repeatability $\pm0.1$mm). The 3D scanner was utilized to acquire 3D information of the skull defect and to export a refined mesh (Fig. 1, bottom, left). We modified the KUKA robotic arm into a cutting workstation by attaching a spindle tool to the robot's end-effector (Fig. 1, bottom right). 

Our intraoperative CCI modification method includes four steps (Fig. 1, middle):
\textbf{1)} The information of a defected skull is collected using a 3D scanner.
\textbf{2)} The scanned data is registered to the CT model space.
\textbf{3)} A cutting toolpath is generated by extracting the defect contour.
\textbf{4)} A cutting robot resizes the CCI according to the generated toolpath.


\begin{figure}[t!]
\centerline{\includegraphics[width=1.0\linewidth]{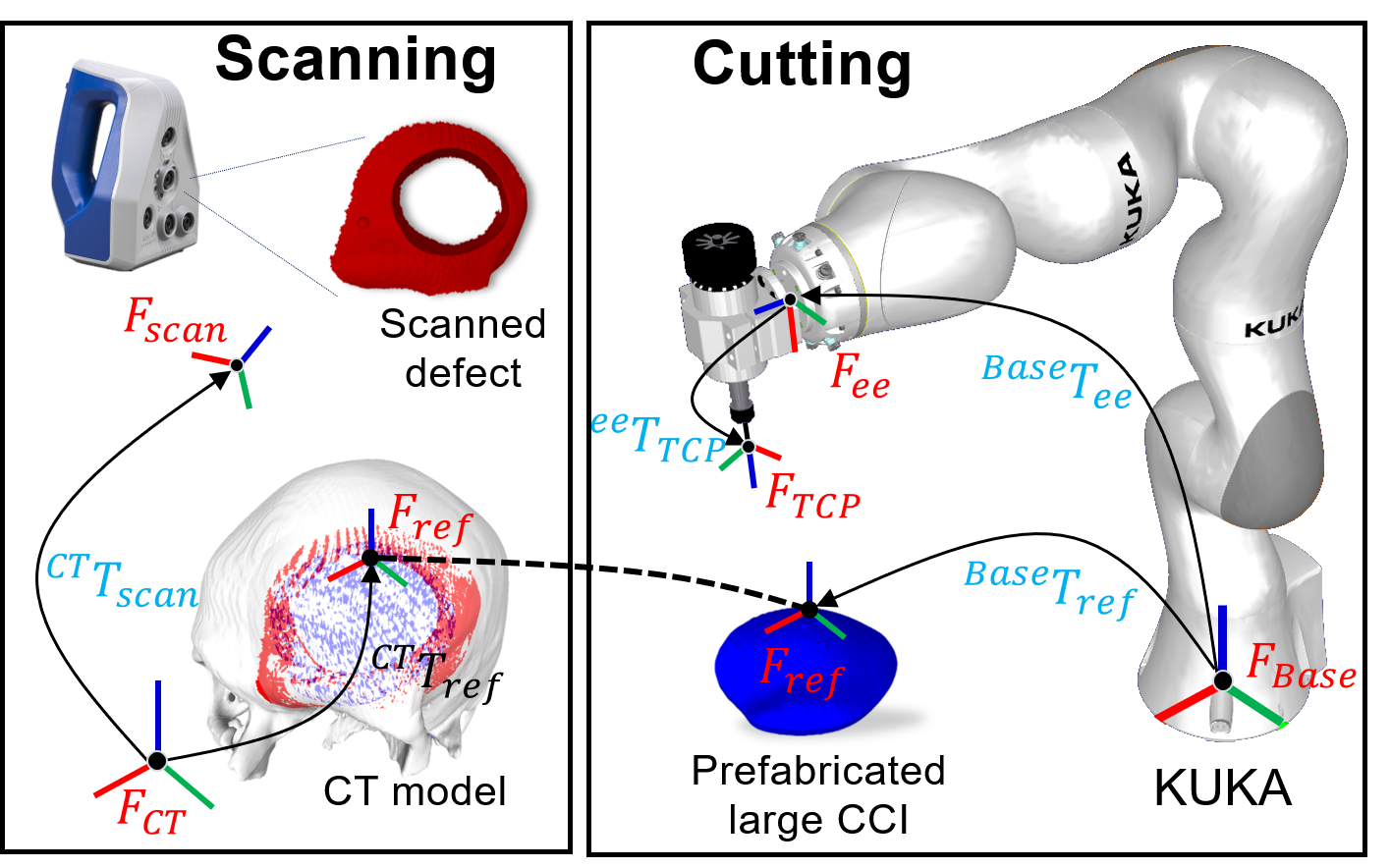}}
\caption{Coordinate transformations between different models. Left: patient-to-CT registration aligns the 3d-scanned defect mesh model (red) $F_{scan}$ to the CT model (white) $F_{CT}$. Right: the robot locates a prefabricated oversized CCI (blue) by finding the reference frame $F_{ref}$ defined by three spherical markers that originally defined in $F_{CT}$. $F_{Base}$ is the robot’s base frame. $F_{ee}$ is the robot’s end-effector frame. $F_{TCP}$ is the calibrated TCP frame. The transformation between different coordinate frames are shown as $T$.}
\label{fig-coordinate-transformation}
\end{figure}


\subsection{3D reconstruction of a patient's defect skull}

A 3D scanning process was first utilized to acquire intraoperative patient data. During this process, the 3D scanner was held by hand at an approximate half meter away from the skull and moved slowly around it. This process could be terminated when there were sufficient 3D data shown in the visualization software (Artec Studio). This process usually takes less than two minutes.

\subsection{Patient-CT registration}
The 3D-scanned data was then registered to the preoperative CT model of the patient's skull. This process transformed the scanned data to the CT model space, as shown in Fig. 2. An iterative closest points (ICP) registration method \cite{besl1992method} was applied to refine the registration. Three anatomical points were artificially designed on the defect skull and were picked from the 3d-scanned model for ICP initialization.

Since the preoperative CT model has two layers separated by the bone thickness, during the registration process, the 3d-scanned data tends to mistakenly overlap with the inner layer of the skull, because of the similar geometric feature between the inner and outer layer. To prevent this problem, we designed a preprocessing algorithm to convert the closed CT model to an open surface mesh by removing the inner layer.

In this method, we defined the 3D position of each vertex in the CT skull mesh as $\mathbf{q}_{i} \in \mathbb{R}^{3}$. The center of the mesh $\mathbf{o} \in \mathbb{R}^{3}$ can be approximated by:
$\mathbf{o} = \frac{\sum_{i = 1}^{n}\mathbf{q}_{i}}{n}$. Then we constructed vector $\mathbf{v}_{i}$, which points from the center $\mathbf{o}$ to each vertex $\mathbf{q}_{i}$. Since each vertex in the skull mesh is also associated with a normal vector $\mathbf{n}_{i}$ of its own. The vectors of the inner layer point to the hollowed space inside the skull towards the center $\mathbf{o}$, while the vectors of the outer layer point to the opposite directions. As a result, the sign of $\mathbf{q}_{i} \cdot \mathbf{n}_{i}$ determines whether this vertex is located in the inner layer or the outer layer. Vertices with negative dot products were removed to keep only the outer layer of the CT mesh model.


\begin{figure}[t!]
\centerline{\includegraphics[width=1.0\linewidth]{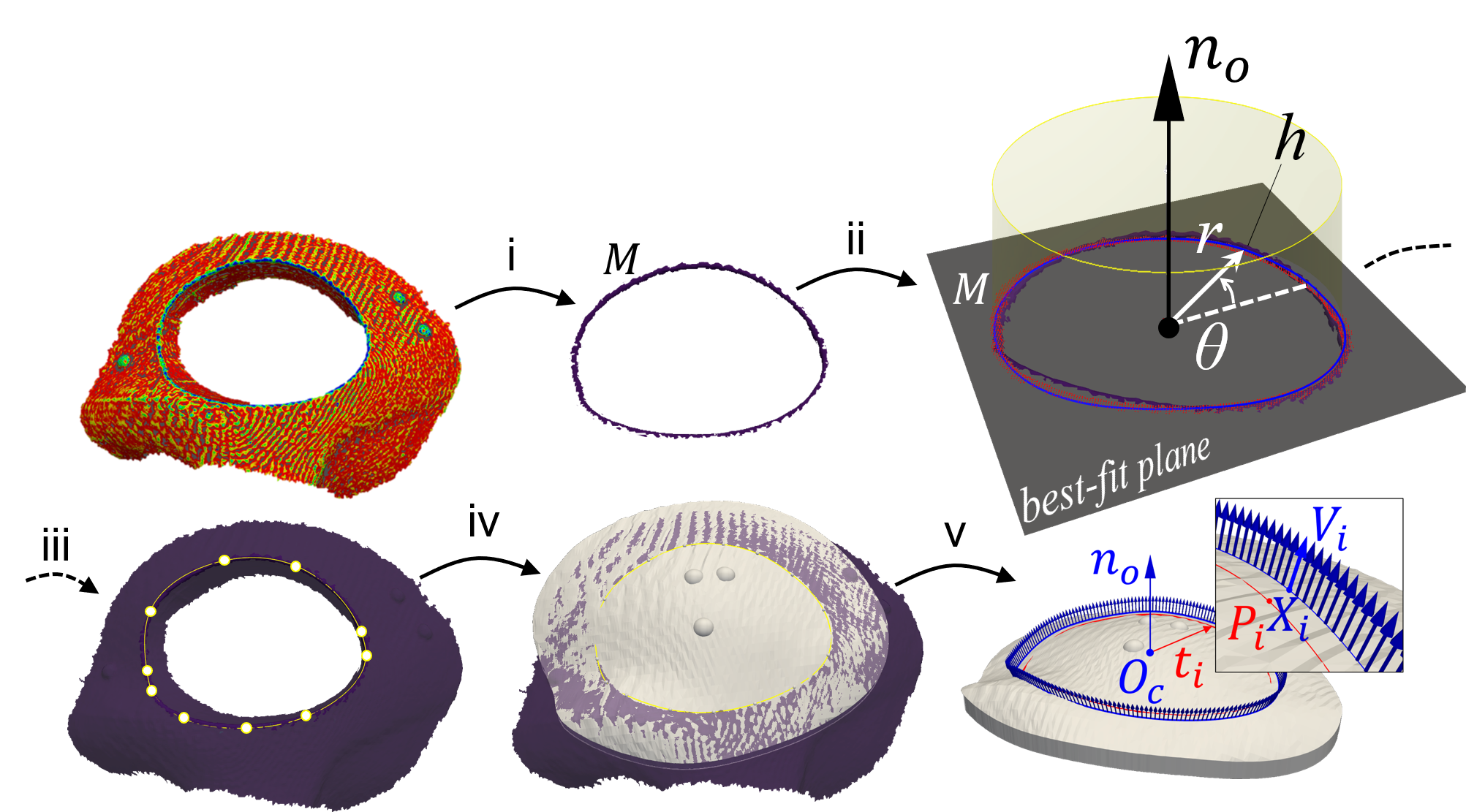}}
\caption{Toolpath generation. (i) A curvature filter followed by manual cleaning is applied to the 3d-scanned defect data to extract the vertices of the defect contour. (ii) The remaining vertices are fitted to a plane, transformed to a local cylindrical coordinate system defined on the plane parameterized as $(\theta,\ r,\ h)$, and then fitted to a polynomial curve. (iii) The fitted curve is converted into a spline curve interpolated through control points. (iv) The control points are projected onto the registered implant's top surface. (v) A cutting toolpath is generated from the spline curve.}
\label{fig-toolpath-generation}
\end{figure}


\subsection{Toolpath Generation}
To generate a toolpath for the subsequent resizing process, the defect contour was first extracted from the 3d-scanned mesh of the skull defect. Then, the 3d-scanned defect was registered to the preoperative CT model. Therefore, the implant was aligned to the 3d-scanned defect model in the CT coordinates. A toolpath consisting of cutting positions and vectors along the extracted contour was generated in the implant coordinate system, as shown in Fig. 3.
We implemented the following steps to generate 3D cutting toolpaths with visualization using Pyvista \cite{sullivan2019pyvista}:

\subsubsection{Curvature Filter}
To extract the contour of the defect, a curvature filter was applied to the vertices of the 3d-scanned mesh and followed by manual cleaning. We utilized a curvature filter to determine the local mean curvature along the surface of the defect and extracted the high curvature value above a designed threshold. The mean curvature $H$ was calculated as  \( H = \frac{1}{2} \left( \kappa_{1}+\kappa_{2} \right)  \) , where  \( \kappa_{1} \) and  \( ~\kappa_{2} \) are the maximum and minimum values of the principal curvature on the mesh \cite{meyer2003discrete}. This filter was able to identify crease changes by the curvature of the surface mesh. After curvature filtering, only the vertices near the dropping edges were kept (Fig. 3, i). Additional manual cleaning was performed to further remove potential redundant vertices that were far off the edge.

\subsubsection{Curve Fitting}
After removing all the redundant vertices, a group of vertices around the defect edge were remained denoted as  \( M \) (Fig. 3, ii). We defined a local cylindrical coordinate system on a best-fit plane of the extracted contour parameterized as $(\theta,\ r,\ h)$. The center of the cylindrical coordinate system was obtained by the mean coordinates of the remaining vertices. A nonlinear least-squares method was then used to fit a closed polynomial curve to these vertices expressed in the cylindrical coordinates.

\subsubsection{Spline Projection}
The fitted curve was then converted into a spline curve, which consists of several control points along the curve. Although the 3d-scanned defect was registered to the CT model after patient-to-CT registration, the extracted defect contour may not perfectly align to the implant's top surface due to registration error. To eliminate the error, the control points were projected onto the top surface of the implant mesh so that the spline curve would adjust its shape to match the curvature of the implant's top surface (Fig. 3, iv).


\begin{figure}[t!]
\centerline{\includegraphics[width=0.9\linewidth]{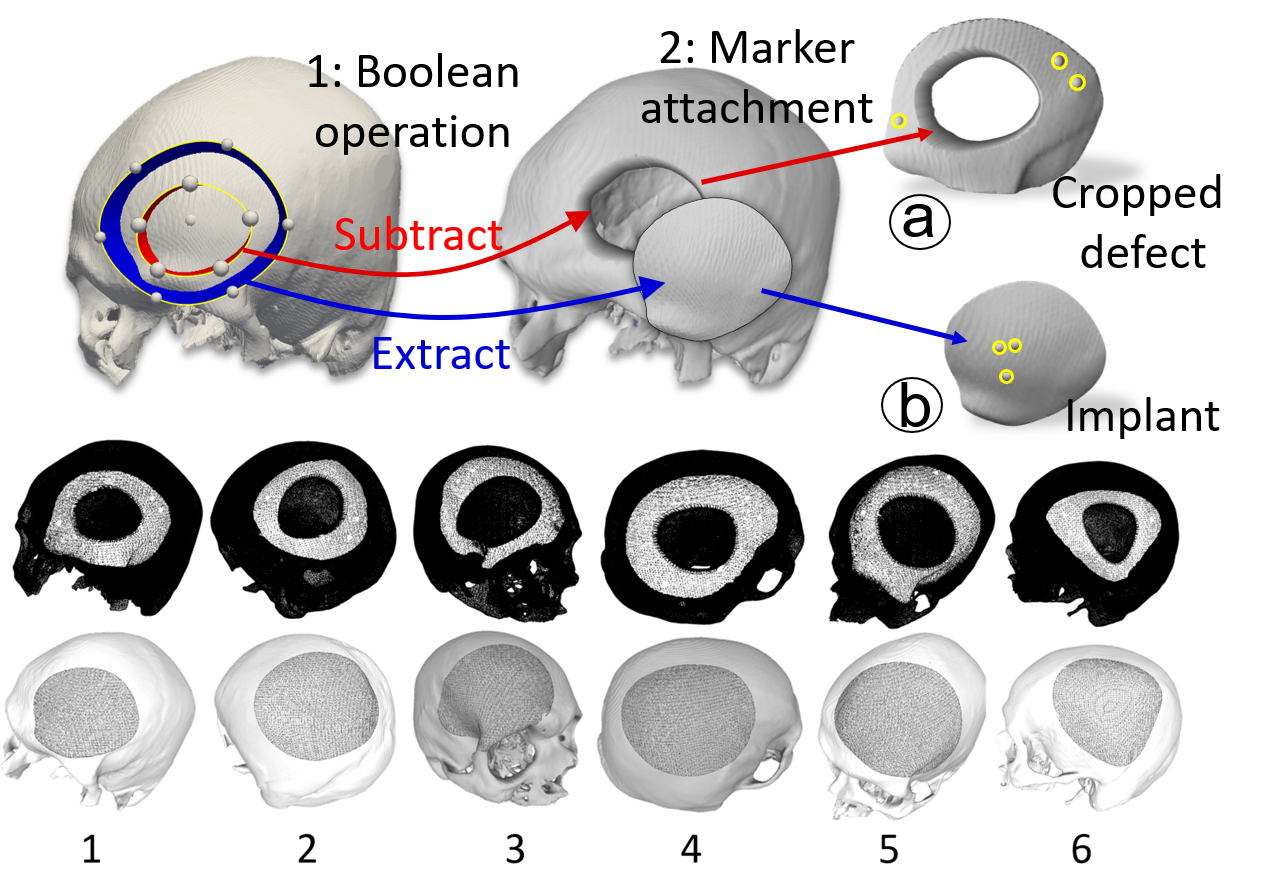}}
\caption{Implant and skull defect generation. Top: the skull defect (a) and the implant (b) are generated by (1) a boolean operation between the skull and two customized contours (red contour: skull defect, blue contour: implant) and (2) attaching spherical markers to their surfaces. The skull defect is cropped to a 3-D printable size. Bottom: the skull defects (top) and implants (bottom) for six different specimens are generated.}
\label{fig-data-creation}
\end{figure}


\subsubsection{Toolpath Generation}
For each discretized point $\mathbf{P_{i}}$ of the spline curve, a unit 3D vector $\mathbf{V_{i}}$ was added to define the tool center point (TCP)'s axis orientation. Each $\mathbf{V_{i}}$ was computed by tilting a constant angle from $\mathbf{n_{o}}$, the normal vector of the best-fit plane, toward $\mathbf{t_{i}}$, a vector defined from the center point $O_{c}$ to each curve point $\mathbf{P_{i}}$. A cutting toolpath was then obtained by combining each curve point $\mathbf{P_{i}}$ with its associated cutting vector $\mathbf{V_{i}}$. To compensate the tool radius, the curve points $\sum \mathbf{P_{i}}$ were expanded to $\sum \mathbf{X_{i}}$ by an offset equal to the radius of the cutting bit (Fig. 3, v).

\section{Experimental Setup}
To evaluate the implant-resizing accuracy of our integrated system, we compared our method with the surgeon's manual-resizing method and an existing optical tracking method. We conducted six experiments with independently generated skull defects with different sizes and shapes. The defect specimens were generated using boolean operations. As shown in Fig. 4 (Top), we first subtracted the mesh inside the red contour from a complete skull to create a defect on the skull. On the same complete skull, the implant mesh inside the blue region was extracted to create its corresponding oversized implant. The defected skull was further cropped by a plane to a 3D printable size and was fabricated using a 3D printer (Stratasys F370, ABS material). Finally, we created three spherical markers on the top surface of each cropped defect and its corresponding implant for Patient-CT registration and implant localization.

\subsection{Method Comparison}
We compared the implants generated by our method with the manual resizing method, as well as the optical tracking method used by Murphy et al. \cite{murphy2015computer}. 
For robotic cutting, the cut depth was set to 3 mm, which is the thickness of the 3d-printed implants. The cut angle was set to 20 degrees for all the generated cutting toolpaths to generate beveled boundaries.

\subsubsection{Manual resizing method}
 We provided the surgeon with pre-designed partial skulls with the generated defects and corresponding oversized implants. The surgeon first outlined the defect contour of each specimen manually on the implant based on his visual judgment. He then resized the implant with a hand-held cutting tool following the outline. 
 
\subsubsection{Optical tracking method}
The optical tracking method used a digitizing instrument and an optical tracking system to trace the defect contour (Fig. 5, a). Instead of manual resizing, the oversized implants were then resized by the same cutting robot and with the same cutting parameters described above.

\subsection{Tool Center Point (TCP) Calibration}
The transformation between the tip of the spindle tool and the robot arm's end-effector was calibrated using a pivot calibration \cite{birkfellner1998calibration}. In this method, we hand-guided the robotic arm to different poses, such that the TCP always touches the tip of a fixed pin. The accuracy of the TCP calibration, measured by the TCP error, is shown in Fig. 5, and the relative pose from $F_{base}$ to $F_{TCP}$:
$$^{Base}T_{TCP} =\ ^{Base}T_{ee} \cdot \ ^{ee}T_{TCP}$$

\subsection{Implant Localization}
The oversized CCI was secured on the robot's working platform with bolts during the resizing process (Fig. 1, bottom, right). In order to obtain the relative position and orientation of the oversized CCI in the robot space (Fig. 2, right), we hand-guided the robot's spindle TCP to touch the tip of each spherical marker separately so that the locations of the reference markers defined in $F_{ref}$ could be transformed to robot's frame $F_{Base}$. The registration between the robot space and CT space was then achieved based on the known locations of the markers, described in the $F_{Base}$ and in the $F_{CT}$ respectively. 

\subsection{Hardware Details}
The integrated system was set up on a computer running Intel Core i7-6820HQ @ 2.7GHz CPU. The 3D scanner (Artec Spider) collects data at 15 HZ. The KUKKA robot is operated using online mode via RoboDK \footnote{RoboDK is an offline programming and simulation software for industrial robots. https://robodk.com/}. The registration between the 3D-scanned model and the CT model was implemented in Meshlab \cite{cignoni2008meshlab}, open-source software for mesh processing. The NDI Polaris optical tracking system operates at 10 Hz (± 0.3 mm tracking accuracy) was used in the comparison experiment.


\begin{figure}[t!]
\centerline{\includegraphics[width=0.9\linewidth]{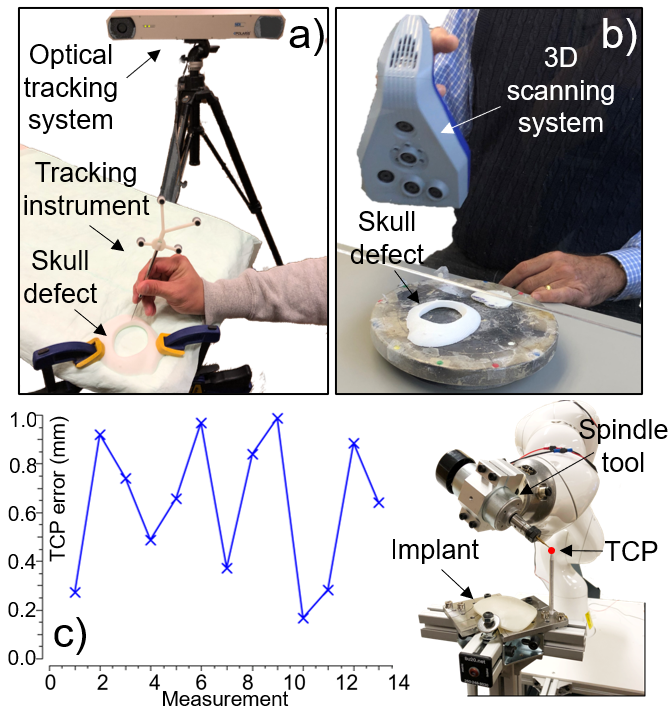}}
\caption{Top: a) collecting the defect contour by optical tracking system. b) scanning the defect contour with a handheld 3D scanner. c) robot TCP calibration errors}
\label{fig-TCP}
\end{figure}


\section{Result}

\subsection{Registration}
1. Optical Tracking Method \newline
Three anatomical markers were artificially added to the original CT models and were 3D printed with the defect specimens. The anatomical points on the printed specimens were localized in the optical tracking system with a tracking instrument (Fig. 5a). Each defect specimen was registered back to the CT coordinate system using point set registration based on singular value decomposition \cite{arun1987least}. The registration error was given by the mean Cartesian distance between the registered anatomical points and the original anatomical points defined in the CT space (Table \ref{tab-registration-errors}).

2. 3D Scanning Method \newline
After 3D scanning the defect specimen, we first manually aligned the three anatomical points on the defect with the original anatomical points defined in the CT model as an initialization. Then ICP was used to fine-tune the registration of the scanned specimen to the original CT model. The error was evaluated by calculating the mean distance between all of the valid vertices and their closest vertices in the original CT model (Table \ref{tab-registration-errors}).

\begin{table}[htbp]
\caption{Registration Errors of Two Methods}
\centering
\resizebox{\linewidth}{!}{
\begin{tabular}{|c|c|c|c|c|c|c|}
\hline
Specimen &
1 & 2 &3 & 4 & 5 & 6\\
\hline
\makecell{Optical Tracking\\ Point Cloud\\ registration (mm)} & 0.39 & 0.38 & 0.32 & 0.32 & 0.30 & 0.34\\
\hline
\makecell{3D Scanning\\ ICP (mm)} & 0.04 & 0.02 & 0.04 & 0.04 & 0.05 & 0.03\\
\hline
\end{tabular}
}
\label{tab-registration-errors}
\end{table}


\begin{figure}[t!]
\centerline{\includegraphics[width=0.95\linewidth]{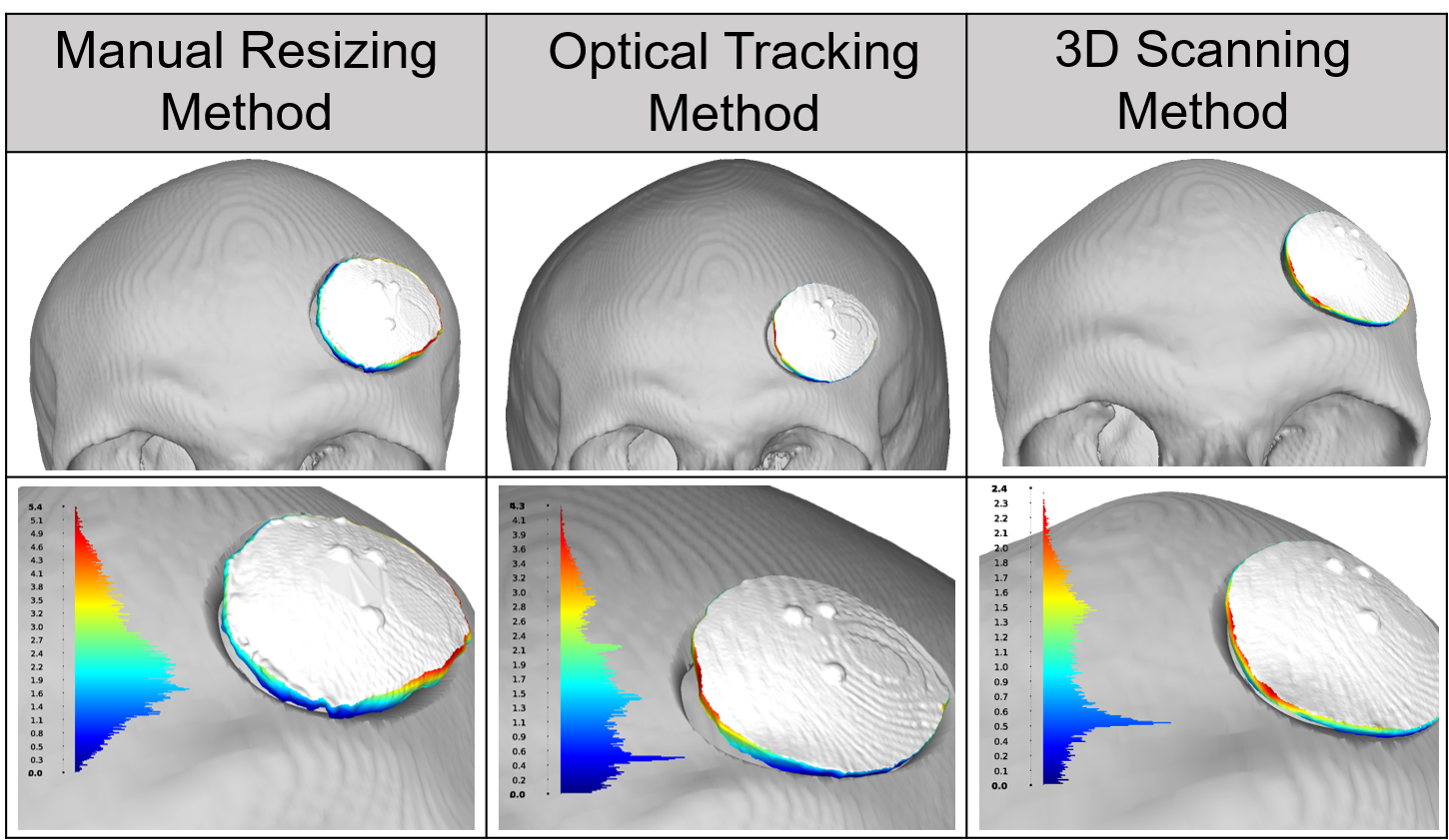}}
\caption{An example of gap distance analysis (specimen 1). Left, middle, right show the results of the conventional manual modification, the existing optical tracking method, and our proposed 3D scanning method, respectively. Top and bottom show their overviews and zoomed views. The color bars in the bottom plots show the gap distance between the implant's boundary and the defect edge.}
\label{fig-dist-analysis}
\end{figure}

\subsection{Evaluation of resized implants}

The post-completed implants were physically fitted to the defect specimens. First, we scanned the implants sitting on their respective defects with a 3D scanner and registered to their original CT models using the artificial anatomical points on the defect specimens. We then 3d-scanned the post-completed implants individually and registered to the previous 3D scans using the three artificial fiducial points created on each implant so that each post-completed implant model could be positioned correctly relative to the ground truth defect model.

We then evaluated the gap distances between the boundaries of the post-completed implants and their corresponding defect edges. The gap distances were visualized in Meshlab \cite{cignoni2008meshlab} (Fig. 6). We used the gap distance distribution (maximum, mean, and standard deviation) to quantify the error for each method for the 6 specimens. Fig. 7 shows the analysis of the max gap distance. For the manual resizing method, all the resized implants were smaller than the defect edges since the surgeon would repeatedly trim the implant until it fits into the defect. Yet, for optical tracking and 3D scanning methods, the post-completed implant could be slightly larger than the defect boundaries, which would not completely fit into the defect.

Among the six defect specimens, the third specimen was considered the most difficult case due to the complex shape above the eye orbit. Although the gap analysis for our proposed approach showed the fourth and the sixth specimens had larger maximum bone gaps than the third specimen, this corresponds to the fact that the resized implants did not completely fit into the defect and would require slight trimming. Among the mentioned three methods, Fig. 8 shows that our proposed 3D scanning method with robotic integration was the only one with the mean gap distance below 1.5 mm.

\subsection{Time cost}
The resizing process by our proposed method took about 10-15 minutes, including the setup time, which was similar to the optical tracking method, whereas the conventional manual approach takes a range from ten to eighty minutes by expert surgeons.


\begin{figure}[t!]
\centerline{\includegraphics[width=0.95\linewidth]{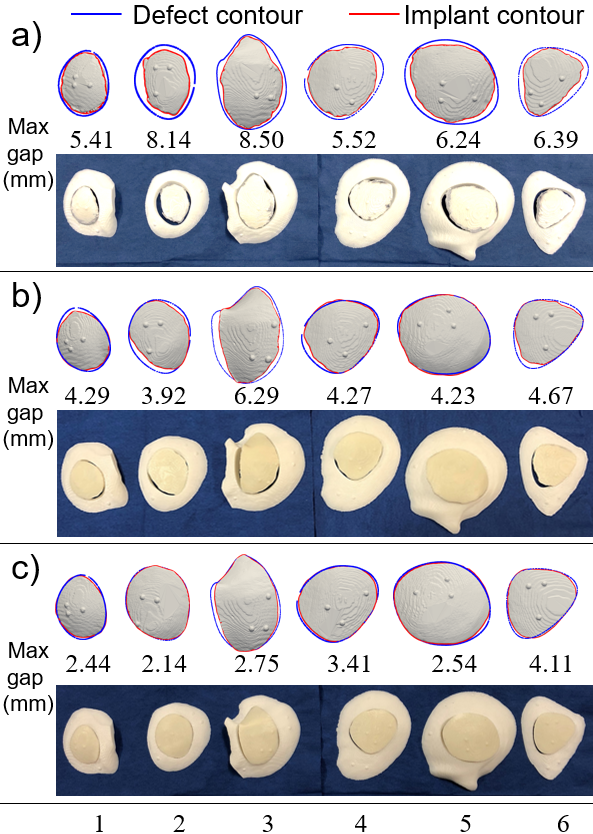}}
\caption{Cutting performance. Visualization of the defect contour (blue curve) and implant contour (red curve) for a) conventional manual modification, b) optical tracking method, and c) 3D scanning method. The numbers in the middle of each plot show the maximum gap distance.}
\label{fig-maximum-gap}
\end{figure}



\begin{figure}[t!]
\centerline{\includegraphics[width=0.9\linewidth]{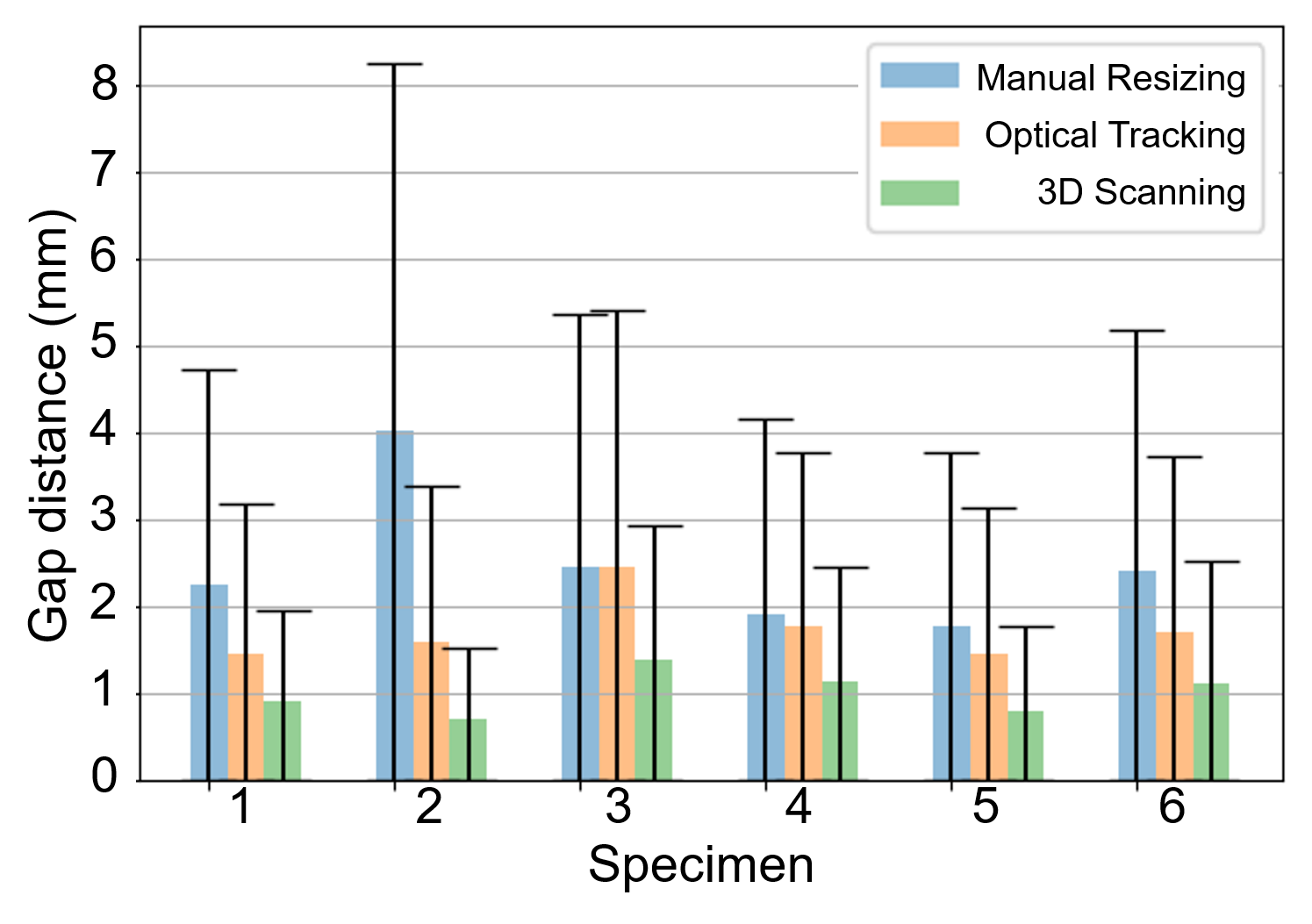}}
\caption{Cutting accuracy evaluation. Mean and standard deviation of the gap distance between the implant and the defect of conventional manual modification (blue), optical tracking method (orange), and 3D scanning method (green) for six specimens.}
\label{fig-final-evaluation}
\end{figure}



\section{Discussion}

We present a novel method for intraoperatively fabricating precise CCI in single-stage cranioplasty. In the proposed method, we first scan the defect to create a mesh model. The mesh model is then registered to the reconstructed 3D model from the preoperative CT in order to define the contour of the defect. Next, a cutting toolpath is generated using a discrete spline curve to represent the defect contour. After localizing the oversized CCI with respect to the robot's base frame, the cutting robot automatically resizes the implant to generate the final shape to fit the defect.

The proposed method improves the accuracy of the cut by 56\% compared to the surgeon’s cut and 42\% compared to the optical tracking method. Moreover, the implant cut boundaries as created by the robot were considerably smoother than those created by the expert surgeon. The smooth boundary may contribute to the better fit of implants to the defect area in actual surgical scenarios. Our proposed method significantly reduced the operation time compared to the expert surgeon's performance time of 10-80 minutes, as reported by Berli et al \cite{berli2015immediate}. The robotic modification of oversized CCIs was shown to be more consistent and accurate when compared to the expert surgeon's performance. Of note, we used an available seven DOF Kuka robot to perform the cutting tasks. However, a cheaper six DOF robot or a five-axis laser cutting machine (e.g. \cite{liu2017design}) can also successfully perform smooth cutting as proposed for this research.

Some of the limitations of the current study are as follows: 1) During the toolpath generation process, the cut angle defining the tool axis attached to the discretized points along the defect contour was constant. In actual surgical scenarios, This may cause problems in fitting the implant if the defect boundary is not beveled uniformly. The extension of this work will include the development of an algorithm that can extract the bevel angle of the defect wall from the scan data. 2) In this study, the manually-tuned, experimentally-determined cutting speed and spinning rate of the tool were not optimized. Additional experiments are needed to evaluate the optimal cutting parameters for smooth cutting of the implant. 3) In the clinical setting, due to the minimally exposed surgical area (Figure 1, Top, iii), the process of registering a patient's defect scan to the CT model may be challenging. A possible remedy is to use two separate 3D scans at different times during the procedure. Prior to the draping of the patient, fiducial marks will be attached to the patient. The first scan will acquire the full exposed head with fiducial marks attached and register this head model to the preoperative CT scan of the patient's skull. After draping the surgical area and subsequent skull resection, a second scan containing the defect area information will be registered to the first scan using the information obtained from the fiducial marks. Thereby, the defect scan can be mapped to the CT model through the intermediate scan obtained prior to draping.

\bibliographystyle{IEEEtran}
\bibliography{bibliography.bib}
\end{document}